\documentclass[letterpaper]{article} 
\usepackage{aaai25}  
\usepackage{times}  
\usepackage{helvet}  
\usepackage{courier}  
\usepackage[hyphens]{url}  
\usepackage{graphicx} 
\usepackage{natbib}  
\usepackage{caption} 
\usepackage{algorithm}
\usepackage{algorithmic}

\usepackage{booktabs}
\usepackage{amsmath} 
\usepackage{color}
\usepackage{multirow}
\usepackage{subfigure}
\usepackage{pifont}

\usepackage{newfloat}
\usepackage{listings}
\DeclareCaptionStyle{ruled}{labelfont=normalfont,labelsep=colon,strut=off} 
\floatstyle{ruled}
\newfloat{listing}{tb}{lst}{}
\floatname{listing}{Listing}
\nocopyright 

\title{AAAI Press Formatting Instructions \\for Authors Using \LaTeX{} --- A Guide}
\author{
    Written by AAAI Press Staff\textsuperscript{\rm 1}\thanks{With help from the AAAI Publications Committee.}\\
    AAAI Style Contributions by Pater Patel Schneider,
    Sunil Issar,\\
    J. Scott Penberthy,
    George Ferguson,
    Hans Guesgen,
    Francisco Cruz\equalcontrib,
    Marc Pujol-Gonzalez\equalcontrib
}

\title{Accelerating LLM Inference Throughput \\via Asynchronous KV Cache Prefetching}
\author {
    Yanhao Dong\textsuperscript{\rm 1},
    Yubo Miao\textsuperscript{\rm 1},
    Weinan Li\textsuperscript{\rm 1},
    Xiao Zheng\textsuperscript{\rm 1},
    Chao Wang\textsuperscript{\rm 1},
    Jiesheng Wu\textsuperscript{\rm 1},
    Feng Lyu\textsuperscript{\rm 2}
}
\affiliations {
    \textsuperscript{\rm 1}Alibaba Cloud\\
    \textsuperscript{\rm 2}School of Computer Science and Engineering, Central South University, Changsha, China\\
    \{dongyanhao.dyh, miaoyubo.myb, william.lwn, zhengxiao.zx, ruiyi.wc, jiesheng.wu\}@alibaba-inc.com, fenglyu@csu.edu.cn
}

\begin{document}

\maketitle

\begin{abstract}

Large Language Models (LLMs) exhibit pronounced memory-bound characteristics during inference due to High Bandwidth Memory (HBM) bandwidth constraints. In this paper, we propose an L2 Cache-oriented asynchronous KV Cache prefetching method to break through the memory bandwidth bottleneck in LLM inference through computation-load overlap. By strategically scheduling idle memory bandwidth during active computation windows, our method proactively prefetches required KV Cache into GPU L2 cache, enabling high-speed L2 cache hits for subsequent accesses and effectively hiding HBM access latency within computational cycles. Extensive experiments on NVIDIA H20 GPUs demonstrate that the proposed method achieves 2.15× improvement in attention kernel efficiency and up to 1.97× end-to-end throughput enhancement, surpassing state-of-the-art baseline FlashAttention-3. Notably, our solution maintains orthogonality to existing optimization techniques and can be integrated with current inference frameworks, providing a scalable latency-hiding solution for next-generation LLM inference engines.

\end{abstract}

%

\section{Introduction}

Large language models (LLMs) have demonstrated transformative potential across multiple domains, including real-time translation \cite{barrault2023seamlessm4t}, automated content generation \cite{achiam2023gpt}, and personalized recommendation systems \cite{he2023large}. While these models exhibit exceptional capabilities in semantic understanding and generation tasks, their massive parameter scale and computational intensity result in significant latency and cost challenges during inference. To address these challenge, the industry has widely adopted specialized accelerators (e.g., NVIDIA GPUs), whose architectural efficiency in memory bandwidth utilization and parallel computation has emerged as a critical determinant of inference speed and infrastructure economics.

However, the inference process of current LLMs exhibits significant memory-bound characteristics. During inference, although the prefill phase can process the entire input sequence through parallel computation, the decoding phase must generate the target sequence token-by-token due to the autoregressive generation mechanism inherent in Transformer architecture. In this process, each decoding step requires loading the Key-Value Cache (KV Cache) of historical sequences from off-chip High Bandwidth Memory (HBM) into compute unit registers. Due to the inability of HBM bandwidth to meet the throughput demands of modern compute units, these frequent off-chip memory accesses result in substantial data movement latency, which has become the critical bottleneck limiting LLM inference throughput.

To address memory-bound bottlenecks, existing optimizations primarily focus on reducing memory access overhead. Algorithmic innovations like FlashAttention \cite{dao2022flashattention,dao2023flashattention} employ tiling and kernel fusion strategies to significantly decrease HBM access volume in self-attention mechanisms, while DeepSpeed-inference \cite{aminabadi2022deepspeed} reduces access frequency through operator fusion into unified kernels. However, while these methods effectively reduce HBM access counts, they fail to achieve latency hiding through overlapping compute and memory operations, thereby remaining susceptible to memory bandwidth limitations.

We conduct systematic hardware performance profiling to investigate the potential optimization opportunities in current LLM inference engines. Focusing on the native XFormers backend of the vLLM inference engine \cite{vLLMpaper}, performance analysis reveals that its attention kernel suffers from significant GPU cache misses during the KV Cache loading phase, leading to frequent accesses to high-latency HBM and triggering massive Warp stalls, which severely degrade inference throughput. Notably, the memory bandwidth underutilization observed in this scenario provides critical insights: strategic utilization of idle bandwidth resources could potentially enhance the memory access efficiency of KV Cache.

Leveraging the hardware capabilities of the NVIDIA Hopper architecture, we propose an L2 cache-oriented asynchronous prefetching method for KV Cache, designed to mitigate memory bandwidth constraints in LLM inference. Specifically, the method strategically schedules idle memory bandwidth resources to proactively prefetch required KV Cache into GPU L2 cache during active compute cycles. This enables direct L2 cache hits for high-speed KV Cache loading, effectively hiding HBM access latency within computation phases while preventing data dependency-induced warp stalls.

To validate the performance advantages of our method, we conducted a series of experiments on mainstream open-source LLM models using NVIDIA H20 GPUs, with the state-of-the-art FlashAttention-3 as the baseline. The experimental results demonstrate that: 1) Our approach achieves up to 2.15× improvement in computational efficiency for the native XFormers attention kernel; 2) The acceleration benefits exhibit a positive correlation with sequence length and batch size; 3) Both single-GPU and multi-GPU configurations demonstrate significantly enhanced end-to-end inference throughput compared to native XFormers implementations, while surpassing the baseline FlashAttention-3.

In short, this paper makes the following contributions:
\begin{itemize}
\item Through systematic hardware performance profiling, we reveal critical bottleneck characteristics in vLLM inference engines and uncover abnormal cache hit patterns during KV Cache accesses.
\item We propose a hardware-software co-designed KV Cache asynchronous prefetching method that effectively hides HBM access latency through computation-transmission overlap mechanisms.
\item Evaluations on NVIDIA H20 GPUs demonstrate that our method achieves up to 1.97× end-to-end inference acceleration on mainstream open-source LLMs.
\end{itemize}

\section{Motivation}

\begin{table}[]
\centering
\begin{tabular}{ccc}
\toprule  
Metric & Value\\
\midrule  
Compute Throughput & 23.35\% \\
Memory Throughput & 47.10\%  \\
L1 Cache Hit Rate (Load)& 0.75\% \\
L2 Cache Hit Rate (Load)& 0.06\% \\
Cycles Per Instruction& 27.68 cycles \\
Stall Long Scoreboard & 21.34 cycles\\
\bottomrule 
\end{tabular}
\caption{Hardware performance metrics of the attention kernel in vLLM's XFormers backend collected using a single NVIDIA H20 GPU running the Llama2-7B model with output tokens set to 4096 and batch size of 64.}
\label{tab:motivation}
\end{table}

Extensive studies have demonstrated that GPU-based large language model (LLM) inference systems exhibit pronounced memory-bound characteristics during autoregressive decoding phases \cite{agrawal2024taming,pmlrv202leviathan23a,patel2024splitwise}. The prevailing vLLM framework \cite{vLLMpaper} employs a paged memory management mechanism for non-contiguous block-wise storage of KV Cache (also referred to as KV Block). This design paradigm induces fragmented memory access patterns, thereby exacerbating the memory-bound nature. Through systematic analysis of the attention kernel implementation in vLLM's native XFormers backend, combined with hardware performance profiling data (as shown in Table \ref{tab:motivation}), we identify three critical performance bottlenecks during the inference process: suboptimal GPU resource utilization, poor cache hit rates, and persistent warp stall occurrences.

\subsection{Suboptimal GPU Utilization}

The experimental data reveals that the XFormers kernel achieves merely 23.35\% compute bandwidth utilization and 47.10\% memory bandwidth utilization, indicating notably suboptimal hardware deployment efficiency. This underutilization phenomenon leaves substantial streaming multiprocessors (SMs) in inactive states, fundamentally constraining inference throughput. From a memory-bound perspective, the limited memory bandwidth utilization inherently caps the achievable compute bandwidth, establishing it as the dominant bottleneck for inference throughput. These findings imply that fully exploiting the untapped memory bandwidth could unlock significant potential for throughput improvement.
\subsection{Poor Cache Hit Rate}

In the GPU memory hierarchy, data load operations follow a tiered cache lookup mechanism: first probing the L1 cache, then descending to the L2 cache upon miss, and ultimately triggering high-latency off-chip High Bandwidth Memory (HBM) access if both caches miss. Taking the NVIDIA H100 GPU as an example, its three-tier storage system (L1/L2/HBM) exhibits a significant performance gap in bandwidth (33 TB/s vs. 12 TB/s vs. 3.35 TB/s). As demonstrated in Table \ref{tab:motivation}, the cache access pattern of XFormers kernel shows significant anomalies: the L1 cache hit rate is merely 0.68\%, and the L2 hit rate drops further to 0.12\%, indicating a near-complete failure of the caching mechanism. This forces all memory requests from SMs to rely entirely on the HBM layer, significantly amplifying data access latency.

\subsection{Persistent Warp Stall}

\begin{figure}
\centering
\includegraphics[scale=0.39]{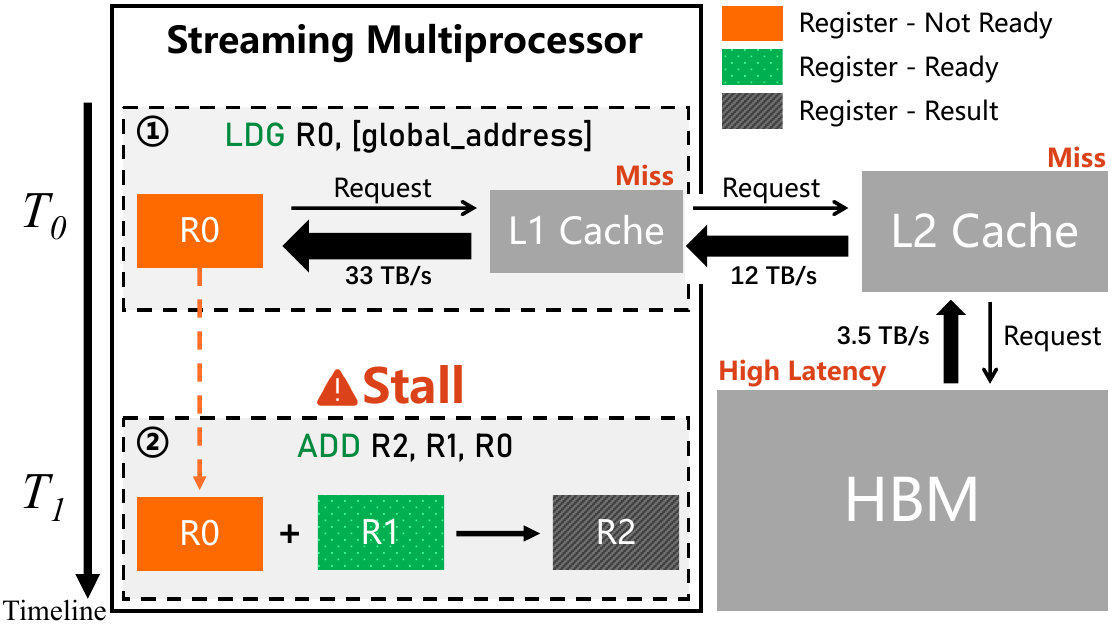}
\caption{Schematic illustration of Stall Long Scoreboard event in NVIDIA GPU.}

\label{warpstall}
\end{figure}

In the NVIDIA GPU architecture, SMs utilize warps as fundamental scheduling units. The XFormers kernel exhibits severe warp stalling phenomena: the average Cycles Per Instruction (CPI) reaches 27.68, with 21.34 cycles (77\%) consumed by "Stall Long Scoreboard" events, forming an absolute performance bottleneck. As illustrated in Figure 1, this stall event occurs when a warp executes load instructions (LDG) to transfer data into registers and encounters cache misses — triggering high-latency HBM accesses — its subsequent arithmetic instructions (e.g., ADD) become non-executable due to register data dependencies, forcing the warp into a stall state until HBM data reaches SM registers. The kernel's near-zero cache hit rate necessitates virtually all memory requests to access HBM, inducing persistent Stall Long Scoreboard events that catastrophically waste GPU computational cycles. Source code analysis reveals that these stall events predominantly cluster during the KV block loading phase, fundamentally rooted in the KV Cache's access pattern exhibiting severe mismatch with cache locality principles.

In summary, the GPU cache misses encountered by the XFormers kernel during KV Block loading directly trigger extensive warp stall cycles, resulting in low computational resource utilization. However, the significant presence of idle memory bandwidth in this scenario reveals a potential optimization pathway: By strategically leveraging underutilized bandwidth resources to systematically optimize the memory access efficiency of KV Cache, the stall cycles caused by HBM's high-latency accesses could be remapped into productive computation cycles. This approach holds promise for breaking through memory-bound constraints and achieving step-change improvements in end-to-end inference throughput.

\section{The Proposed Method}

\begin{figure*}[h]
	\centering
    \subfigure[Native XFormers]{
    	\begin{minipage}[b]{0.48\textwidth}
   		\includegraphics[width=1\textwidth]{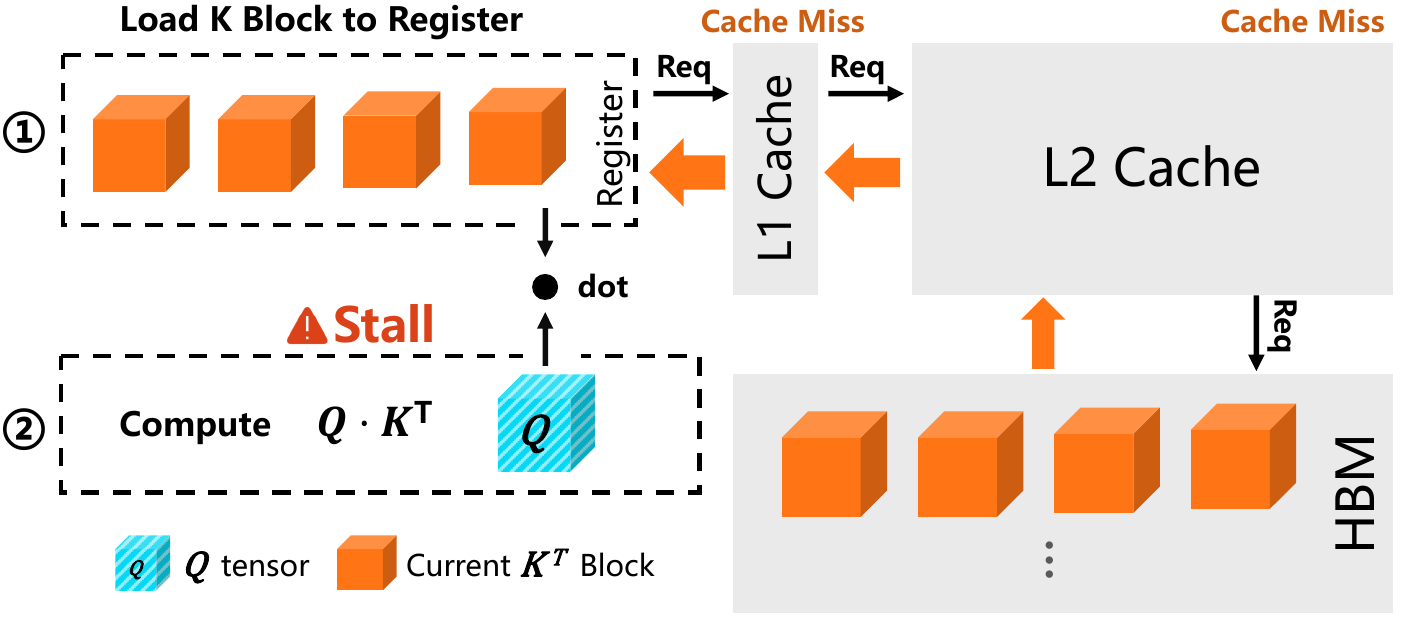}
    	\end{minipage}
	\label{fig:xformersoverview}
    }
    \subfigure[The proposed method]{
    	\begin{minipage}[b]{0.48\textwidth}
   		\includegraphics[width=1\textwidth]{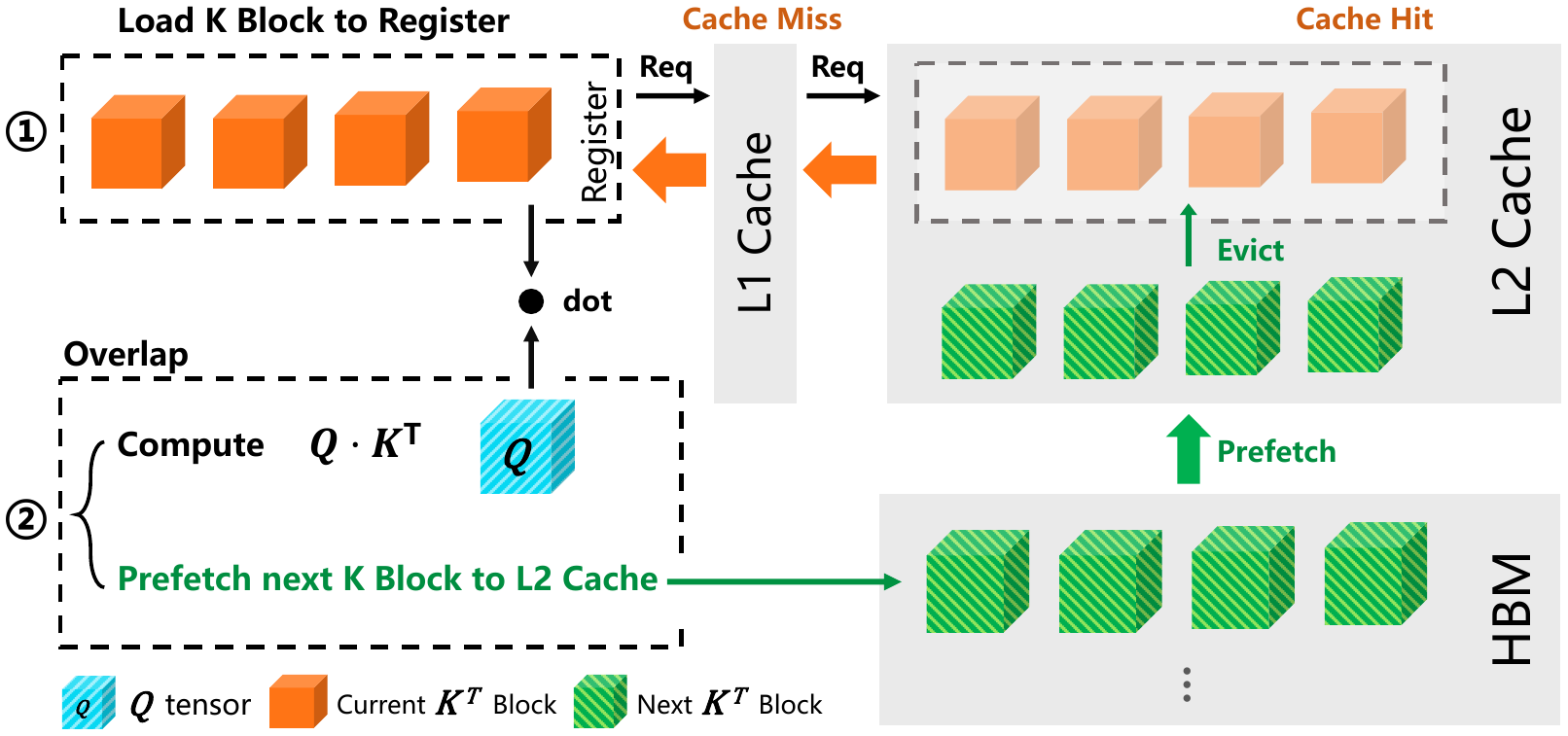}
    	\end{minipage}
	\label{fig:thisworkoverview}
    }
    \caption{$Q \cdot K^T$ computation flow in a single iteration for native XFormers and the proposed method, illustrated with a thread block configuration containing 4 warps.}
\end{figure*}

Building upon the preceding performance analysis, this paper proposes an asynchronous KV Cache prefetching method targeted at XFormers kernel. This method orchestrates idle memory bandwidth resources to proactively prefetch impending KV Blocks into the GPU's L2 cache during compute unit active phases, enabling attention kernels to achieve direct L2 cache hits for low-latency KV Cache acquisition. This design fundamentally circumvents stalling cycles induced by Stall Long Scoreboard events. By overlapping HBM access latency with compute cycles, the proposed method effectively breaks the throughput bottleneck inherent in the native XFormers implementation.

\subsection{Prefetch KV Blocks}

vLLM distributes the KV Cache of each sequence across multiple KV Blocks and locates the physical addresses of these blocks through a block table lookup, effectively reducing memory fragmentation. Specifically, each block exclusively stores KV Cache data for a single attention head and contains a fixed-length token span (typically 16 tokens). This design creates a dual-dimensional distribution pattern in multi-head attention scenarios: the KV Cache of a single token is scattered across $H$ blocks (where $H$ is the number of attention heads), while each block aggregates KV Cache data of the same attention head from multiple contiguous tokens.

Building upon the aforementioned KV Block distribution, the XFormers kernel employs a tailored parallel strategy manifested in two hierarchical levels:
\begin{itemize}
  \item \textbf{Thread Block-Level Parallelism}: A single CUDA thread block (configured as [$N_{thread}$, 1, 1]) acts as the fundamental execution unit, dedicated to processing a single attention head per sequence. Internally, each thread block is divided into multiple warps (32 threads/warp), with each warp handling computations for a single KV Block, enabling a single thread block to concurrently process multiple KV Blocks;
  \item \textbf{Grid-Level Parallelism}: The grid is dimensioned as $[H, B, 1]$ (where $B$ represents the batch size) to schedule multiple thread blocks, achieving parallel processing of multiple attention heads and input sequences.
\end{itemize}

Taking the $Q \cdot K^T$ computation workflow as an example, the native XFormers employs a block-wise iterative loading mechanism during the attention computation phase: each warp independently completes the loading and computation of a K Block within a single iteration until all blocks in the sequence are processed. As shown in Figure \ref{fig:xformersoverview} (using a 4-warp thread block configuration as an example), this execution flow can be divided into two critical stages: \ding{172} Each warp loads a current K Block from global memory into register; 
\ding{173} The warps performs $Q \cdot K^T$ operations using the pre-resident Q tensor in registers with the current K Blocks. Notably, when the current K Blocks miss the cache, the system triggers high-latency HBM access operations, which in turn induce Stall Long Scoreboard events — these events force the corresponding warps into stall state, becoming the core bottleneck limiting kernel performance.

To address the aforementioned bottleneck, this study proposes a hardware-software co-designed KV Block prefetching optimization method, as illustrated in Figure \ref{fig:thisworkoverview}. Using the $Q \cdot K^T$ computation workflow as an example, the core innovation lies in: During the computational cycle when warps execute the current iteration's $Q \cdot K^T$ operations, the method leverages GPU's asynchronous prefetch capability to proactively load the next iteration's required K Blocks from high-latency HBM into L2 Cache. Since the K Blocks for the current iteration have already been loaded into registers, its corresponding L2 Cache lines can be safely evicted without performance degradation. This design ensures that subsequent iterations' K Block requests achieve L2 Cache hits, thereby significantly reducing warp stalls caused by Stall Long Scoreboard events. The detailed algorithmic implementation is presented in Algorithm \ref{alg:algorithm}. The effectiveness of this method hinges on two critical mechanisms: First, the idle memory bandwidth's capacity to efficiently accommodate additional data transfer demands introduced by prefetch operations. Second, concurrent execution of $Q \cdot K^T$ matrix computations and K Block prefetching enables computation-to-memory overlap, effectively masking HBM access delays.

The proposed prefetching method demonstrates equivalent applicability to V Blocks. During a warp's execution of V Block loading operations in the current iteration, the warp concurrently initiates asynchronous prefetching of subsequent iteration’s V Block. This method enables parallel execution of prefetch operations and $logits \cdot V$ computation, achieving latency masking through compute-memory access overlap mechanisms.

\begin{algorithm}[tb]
\caption{Prefetching K Block to L2 Cache.}
\label{alg:algorithm}
\textbf{Input}: Block table: $\mathbf{bt}$; Warps number: $\mathbf{w}$;

\hspace{3em}Block index range: $\mathbf{[s, e)}$.
\begin{algorithmic}[1] 
\STATE Let $block\_idx = \mathbf{s}$;
\WHILE{$block\_idx < \mathbf{e}$}

\STATE Lookup $\mathbf{bt}[block\_idx] \to k\_ptr$; \ \ // {Current Block} 
\STATE Load K Block from  $k\_ptr$ to registers;
\IF {$block\_idx + \mathbf{w} < \mathbf{e}$}
\STATE Lookup $\mathbf{bt}[block\_idx + \mathbf{w}] \to next\_k\_ptr$; \\ // Next Block
\STATE \textcolor[rgb]{0,0.4,0}{Perfetch K Block from $next\_k\_ptr$ to L2 Cache;}
\ENDIF
\STATE Compute $QK^T$ for current K block;
\STATE $block\_idx \gets block\_idx + \mathbf{w}$;
\ENDWHILE
\end{algorithmic}
\end{algorithm}

\subsection{GPU Prefetch Interface}

NVIDIA CUDA introduces an L2 cache-oriented asynchronous prefetch mechanism through the PTX instruction \texttt{cp.async.bulk.prefetch.L2} \cite{nvidiaPTXIntroductionx}. This non-blocking instruction initiates asynchronous data prefetching from specified memory locations into the L2 cache. Notably, this instruction requires hardware support for Compute Capability 9.0 or higher, specifically designed for GPUs based on the Hopper architecture (e.g., H100/H20) and subsequent iterations of GPU devices.

\subsection{Prefetch Benefit vs. L2 Cache Capacity}

\begin{table}[]
    \renewcommand{\arraystretch}{1.3}
    \centering
    \begin{tabular}{|c|c|c|c|c|c|c|}
        \hline
        $b$ & $d_h$ & $T_{block}$ & $N_{thread}$ & $H$ \\
        \hline
        2 & 128 & 16 & 128 & 32 \\
        \hline
    \end{tabular}
    \caption{Typical hyperparameter values for single-GPU inference of Llama2-7B model with FP16 precision.}
    \label{llama2hyper}
\end{table}

The proposed prefetching method relies on residency of KV Blocks within the GPU L2 cache to achieve performance gains. Given that the total size of K/V Blocks required per iteration dynamically varies with configuration hyperparameters (e.g., number of attention heads, batch size), newly prefetched Blocks will evict previously cached valid ones when cumulative prefetch volume exceeds the L2 cache's physical capacity. Such capacity constraints establish a theoretical boundary for the prefetch method's performance enhancement potential. This section quantitatively analyzes the performance improvement limits of the proposed method.

The memory footprint of a single block can be derived through the following computational formula:
\begin{equation}
M_{\text{block}} = b \cdot d_h \cdot T_{block}
\label{singleblock}
\end{equation}
where $ b $ denotes the byte size per model parameter, $ d_h $ represents the dimension of a single attention head, and $ T_{block} $ indicates the number of tokens contained in each block.

Since each K/V Block exclusively contains the K/V Cache of a single attention head, the total memory footprint of Blocks processed in a single iteration is determined by the following formula under the parallelization strategy with thread block dimension [$N_{thread}$, 1, 1] and grid dimension  [$H$, $B$, 1(for the standard multi-head attention (MHA) mechanism)]:

\begin{equation}
M_{\text{total}} = M_{\text{block}} \cdot \frac{N_{thread}}{32} \cdot H \cdot B
\label{totalblock}
\end{equation}

Here, $\frac{N_{thread}}{32}$ represents the number of warps in a thread block. Notably, the total memory footprint of Blocks loaded per iteration is independent of the sequence length; a longer sequence results in more iterations, but the number of Blocks processed per iteration remains fixed.

Taking FP16-precision Llama2-7B model inference on a single GPU as an example, Table \ref{llama2hyper} provides typical hyperparameter values. Assuming a batch size of 1, the memory footprint of a single Block is derived as 4 KB via Equation \ref{singleblock}, while the total data volume of Blocks processed per iteration is calculated as 512 KB using Equation \ref{totalblock}. Users typically employ larger batch sizes to pursue higher throughput. For the NVIDIA H100 GPU, its 60 MB L2 cache theoretically supports the full residency of up to 120 batches of K/V Blocks, forming the theoretical upper bound of performance acceleration for the proposed method. When the practical batch exceeds the upper limit, partially resident blocks still yield marginal performance gains through cache hits. Notably, by adjusting the cache eviction priority \cite{nvidiaPTXEviction} of prefetched data, the cache thrashing caused by capacity overflow can be effectively mitigated.

\section{Experiment}

\begin{table*}[!h]
\centering
\resizebox{\linewidth}{!}{
\begin{tabular}{@{}c|cc|cc|cc|cc@{}}
\toprule
\multirow{2}{*}{Metric} & \multicolumn{2}{c|}{Llama2-7B} & \multicolumn{2}{c|}{Llama3-8B} & \multicolumn{2}{c|}{Qwen2.5-7B} & \multicolumn{2}{c}{Qwen2.5-14B} \\ \cmidrule(l){2-9} 
                               & XFormers & This Work & XFormers & This Work & XFormers & This Work & XFormers & This Work \\ \midrule
Duration (us)                  & 293.47   & 159.14    & 226.53   & 119.90    & 232.03   & 107.71    & 272.19   & 143.94    \\ \midrule
Compute Throughput (\%)        & 23.35    & 48.22     & 29.61    & 61.15     & 25.29    & 60.95     & 30.52    & 63.49     \\ \midrule
Memory Throughput (\%)         & 47.10    & 86.88     & 23.44    & 66.96     & 20.05    & 63.50     & 24.16    & 68.44    \\ \midrule
L2 cache Hit Rate (\%)         & 0.06     & \underline{43.70}     & 38.35    & \underline{73.01}     & 55.90    & \underline{82.66}     & 51.43    & \underline{77.11}     \\ \midrule
Cycles Per Instruction (cycle) & 27.68    & 9.28      & 21.42    & 7.32      & 22.68    & 7.36      & 20.94    & 7.24      \\ \midrule
Stall Long Scoreboard (cycle)  & 21.34    & \underline{4.13}      & 16.66    & \underline{2.05}      & 16.38    & \underline{1.87}      & 16.17    & \underline{1.96}      \\ \midrule
Speedup                 & \multicolumn{2}{c|}{1.84$\times$}     & \multicolumn{2}{c|}{1.89$\times$}     & \multicolumn{2}{c|}{2.15$\times$}      & \multicolumn{2}{c}{1.89$\times$}        \\ \bottomrule
\end{tabular} }
\caption{Attention kernel performance comparison across models using native XFormers backend vs. proposed method. Batch size and output tokens are taken as 64 and 4K, respectively. Models are run on an NVIDIA H20 GPU.}
\label{kernelperf}
\end{table*}

In this section, we evaluate the effectiveness and performance of the proposed methodology. We specifically examine: 1) How proposed prefetching method affects CUDA attention kernel performance 2) How prefetching improves vLLM's end-to-end inference throughput.
\subsection{Experiment Setup}

\subsubsection{Hardware Configuration} The experimental platform is equipped with 4× Intel Xeon Platinum 8469C processors delivering 192 cores and 8× NVIDIA Hopper-architecture H20 GPUs — each H20 GPU contains 60MB L2 cache, 96GB HBM with 4.0TB/s memory bandwidth.

\subsubsection{LLM Model} Our experiments employ community-adopted open-source LLMs — Llama2-7B, Llama3-8B, and Qwen2.5-7B/14B, all in fp16 precision — to evaluate the prefetching method's performance under single-GPU and multi-GPU inference. The critical architectural divergence lies in attention mechanisms: Only Llama2-7B employs standard Multi-Head Attention (MHA, 32 heads) \cite{NIPS2017_3f5ee243}, while others utilize Grouped-Query Attention (GQA) \cite{Ainslie2023GQATG} - Llama3-8B employs 8 KV heads serving 32 Q heads (4:1), Qwen2.5-14B configures 40Q:8KV (5:1), and Qwen2.5-7B adopts an aggressive 28Q:4KV (7:1) arrangement. This architectural variance directly impacts the optimization headroom of our method.

\subsubsection{Baseline \& arguments} we select vLLM v0.7.1's native XFormers backend and the state-of-the-art FlashAttention-3 (FA3) \cite{NEURIPS2024_7ede97c3} backend implementation as baseline. In our experiments, we fixed the input token length at 512, varied the output tokens from 512 to 8192, and adjusted the inference batch size from 1 to 128 to investigate the impact of batch size and sequence length on the performance of the proposed method.

\subsection{Attention Kernel Performance}

We conduct model inference on a single GPU while profiling CUDA kernel performance metrics, specifically evaluating the performance enhancement of our prefetching method over the native XFormers backend for attention kernels, as detailed in Table \ref{kernelperf}. All measurements were captured at fixed sampling points to ensure positional consistency, where L2 cache hit rate specifically measures load operation hits.

Experimental results demonstrate that attention architectures critically influence the cache hit rate in native XFormers: The MHA-based Llama2-7B achieves merely a 0.06\% baseline hit rate, whereas GQA models exhibit progressive improvement with increasing Q/KV head ratios - attaining 38.35\%, 51.43\%, and 55.90\% for Llama3-8B (4:1), Qwen2.5-14B (5:1), and Qwen2.5-7B (7:1) respectively. This discrepancy stems from CUDA grid-level parallel processing of multiple attention heads: In MHA model, each Q head requires independent loading of dedicated KV blocks, while GQA model enable single KV block loads to service multiple Q heads, thereby enhancing cache reuse efficiency. The proposed prefetching method further enhances kernel cache performance by proactively maintaining KV blocks in the L2 cache, increasing the cache hit rate to 43.70\%, 73.01\%, 77.11\%, and 82.66\% across the respective models.

Concurrently, the native XFormers backend exhibits significant warp stalls due to suboptimal cache hit rates – with the Stall Long Scoreboard consuming over 72\% of cycles per instruction (CPI), establishing itself as the dominant performance limiter. The proposed prefetching method effectively mitigates this issue through L2 cache hit rate optimization, demonstrating measurable improvements across four models: In Llama2-7B, LLaMA3-8B, Qwen2.5-7B, and Qwen2.5-14B, scoreboard stall cycles are reduced from baseline 21.34/16.66/16.38/16.17 to 4.13/2.05/1.87/1.96 respectively, driving 65.8\%-67.5\% reductions in attention kernel CPI. Concurrent enhancements in computational resource utilization show substantial gains – peak improvements of 35.66\% in computing throughput and 44.45\% in memory throughput. Ultimately, kernel duration is significantly accelerated, achieving speedup ratios ranging from 1.84× to 2.15×.

\subsection{End-to-end Inference Performance}
\subsubsection{Single-GPU Throughput}
\begin{figure*}
	\centering
    \subfigure[Llama2-7B (32Q:32KV)]{
    	\begin{minipage}[b]{0.23\textwidth}
   		\includegraphics[width=1\textwidth]{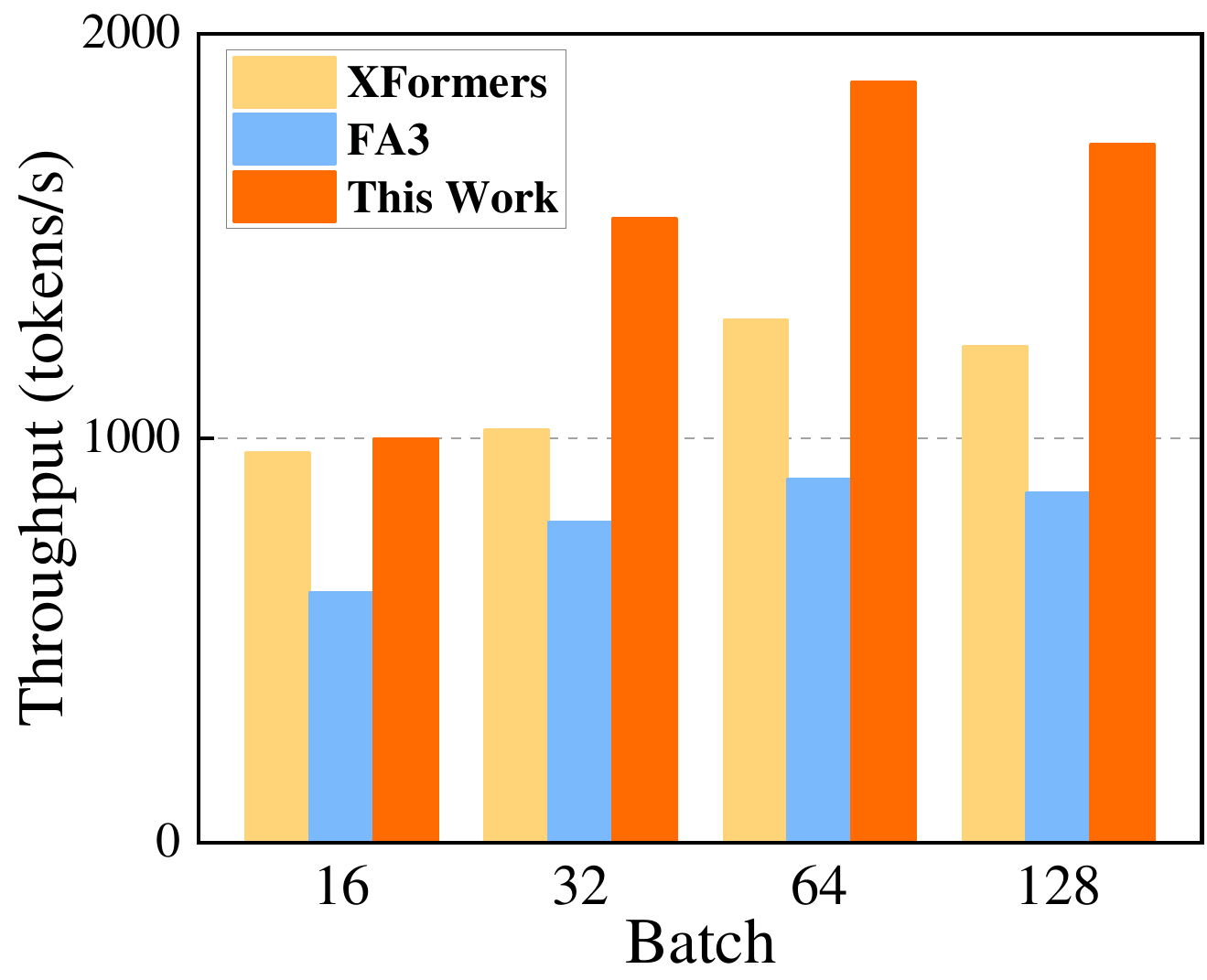}
    	\end{minipage}
    }
    \subfigure[Llama3-8B (32Q:8KV)]{
    	\begin{minipage}[b]{0.23\textwidth}
   		\includegraphics[width=1\textwidth]{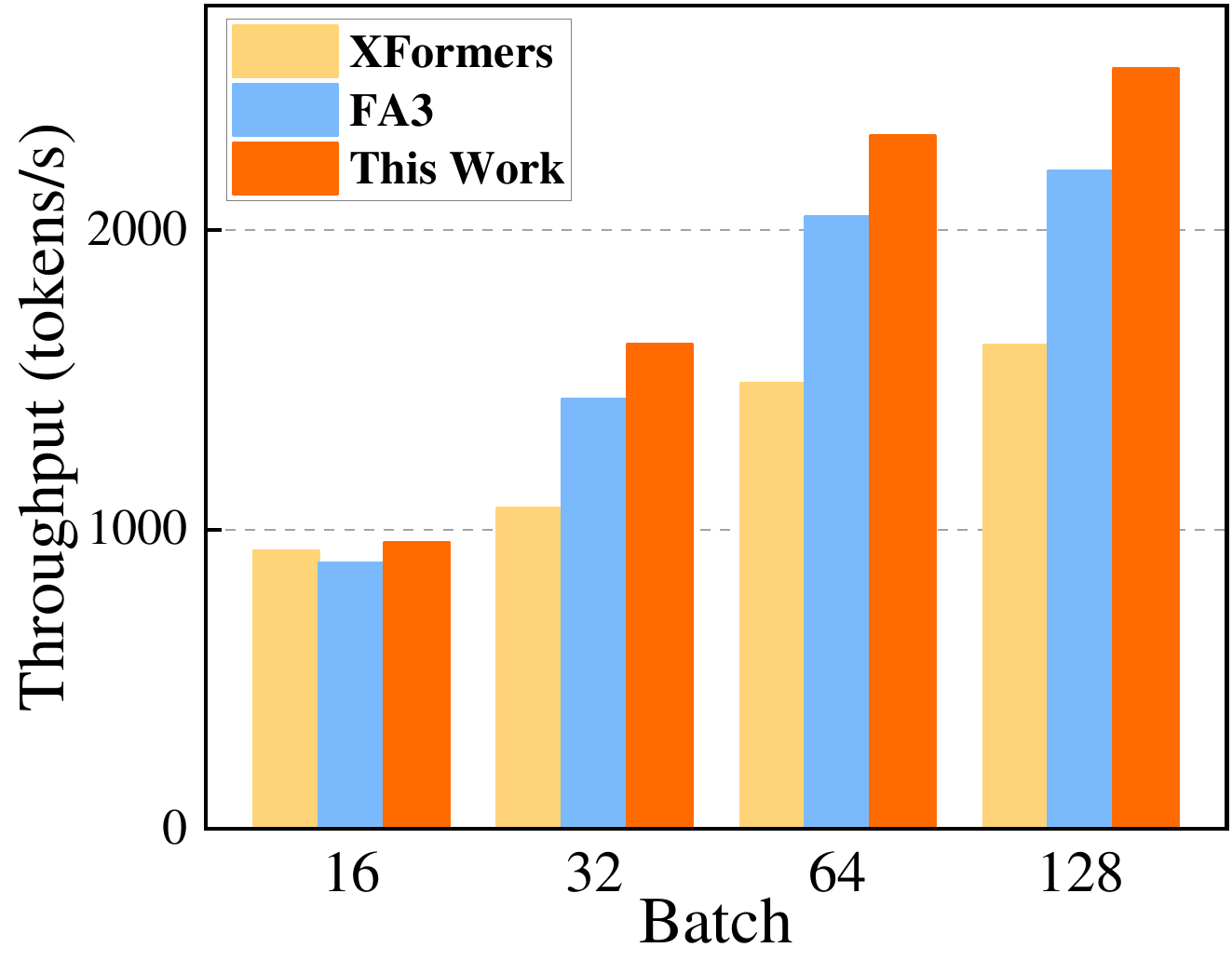}
    	\end{minipage}
    }
    \subfigure[Qwen2.5-7B (28Q:4KV)]{
    	\begin{minipage}[b]{0.23\textwidth}
   		\includegraphics[width=1\textwidth]{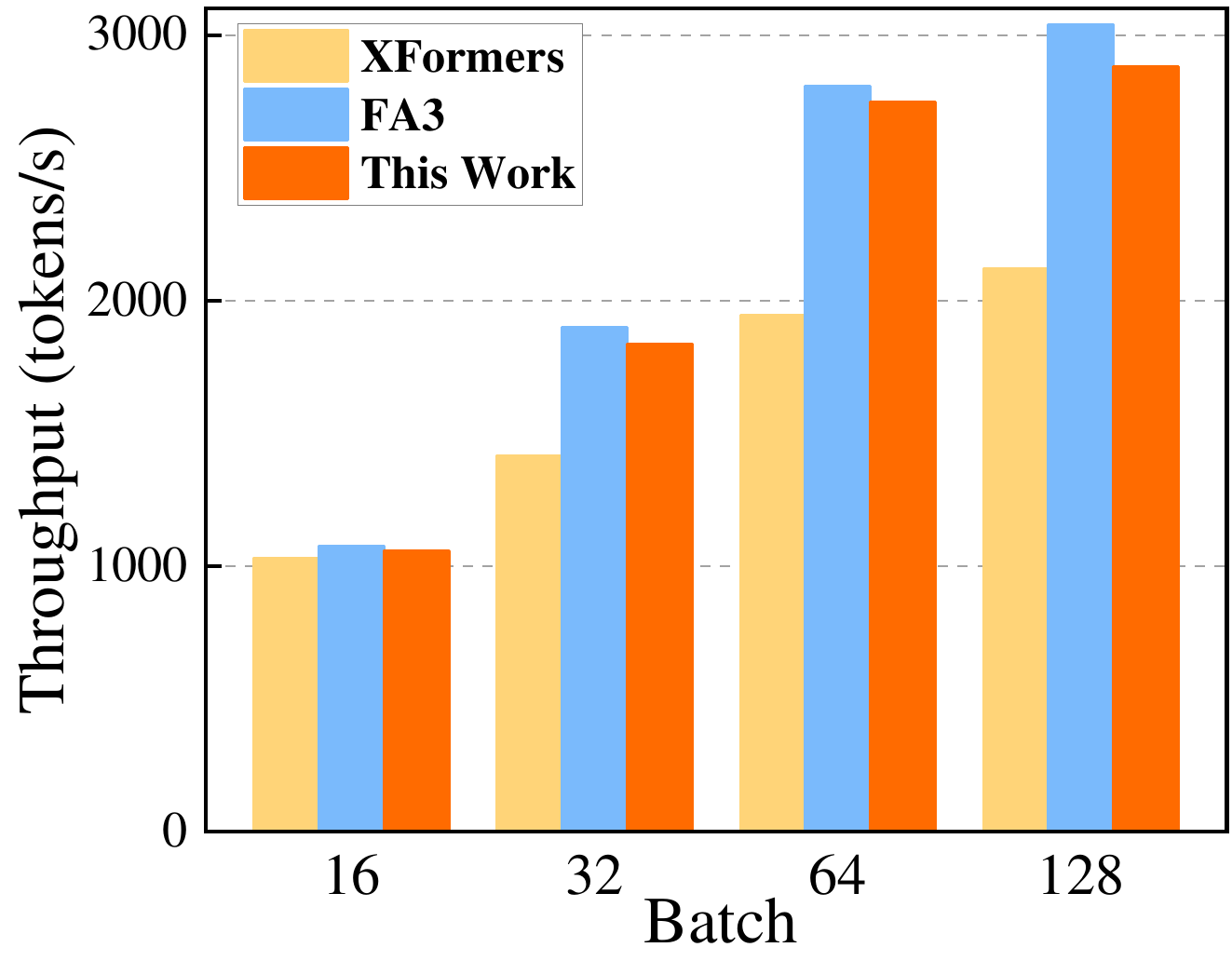}
    	\end{minipage}
    }
    \subfigure[Qwen2.5-14B (40Q:8KV)]{
	\begin{minipage}[b]{0.23\textwidth}
	\includegraphics[width=1\textwidth]{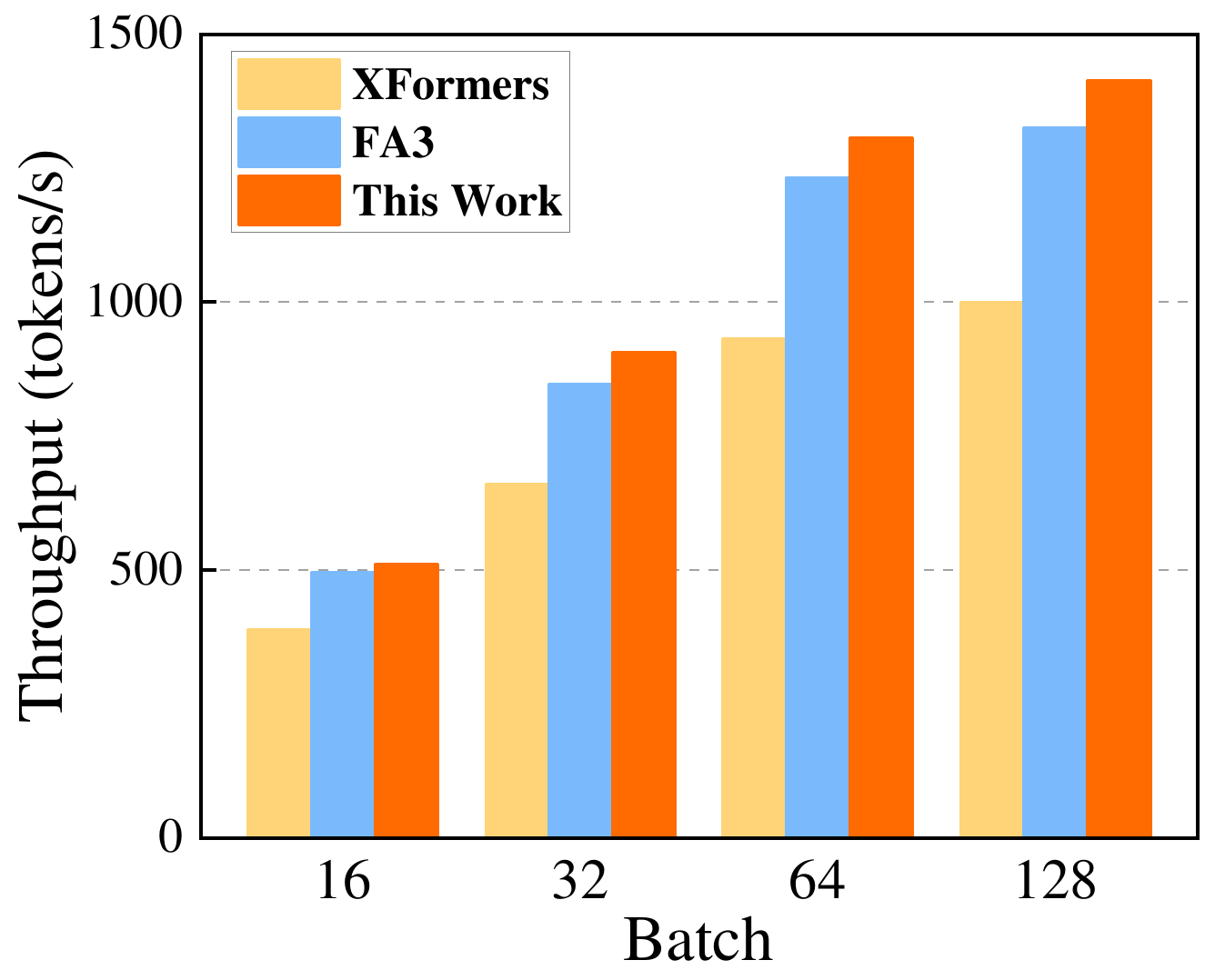}
	\end{minipage}
    }
    \caption{Single-GPU end-to-end inference throughput comparison across backends with fixed 2048 output tokens on H20.}
    \label{fig:singlegpu}
\end{figure*}

We first conducts an end-to-end inference throughput optimization assessment under a single H20 GPU configuration. By executing benchmark tests across multiple models with output tokens fixed at 2048 and batch sizes varying from 16 to 128, Figure \ref{fig:singlegpu} illustrates the end-to-end throughput (tokens/s) data for different models under different backends.

Experimental results demonstrate that prefetching method significantly outperforms the native XFormers implementation across all evaluated models, achieving maximum throughput improvements of 51\%, 57\%, 41\%, and 41\% on Llama2-7B, Llama3-8B, Qwen2.5-7B, and Qwen2.5-14B, respectively. When compared with the state-of-the-art FA3 backend, our approach exhibits consistent performance advantages on all models except Qwen2.5-7B, delivering maximum gains of 110\%, 15\%, and 7\% on the remaining models. Notably, a performance regression of 2-5\% is observed on Qwen2.5-7B versus FA3, which stems from divergent attention architectures: Proposed prefetching optimization accelerates KV Cache loading per attention head, with performance gains being proportional to KV head count. In MHA architectures where each query head accesses independent KV heads, the optimization achieves full benefits. However, GQA architectures share single KV heads across multiple query heads, causing prefetching benefits to decay linearly with KV head reduction. Qwen2.5-7B's aggressive GQA configuration (7:1) severely limits optimization potential due to sparse KV head distribution.

\subsubsection{Batch size \& sequence length}

\begin{figure}
\centering
\includegraphics[scale=0.2]{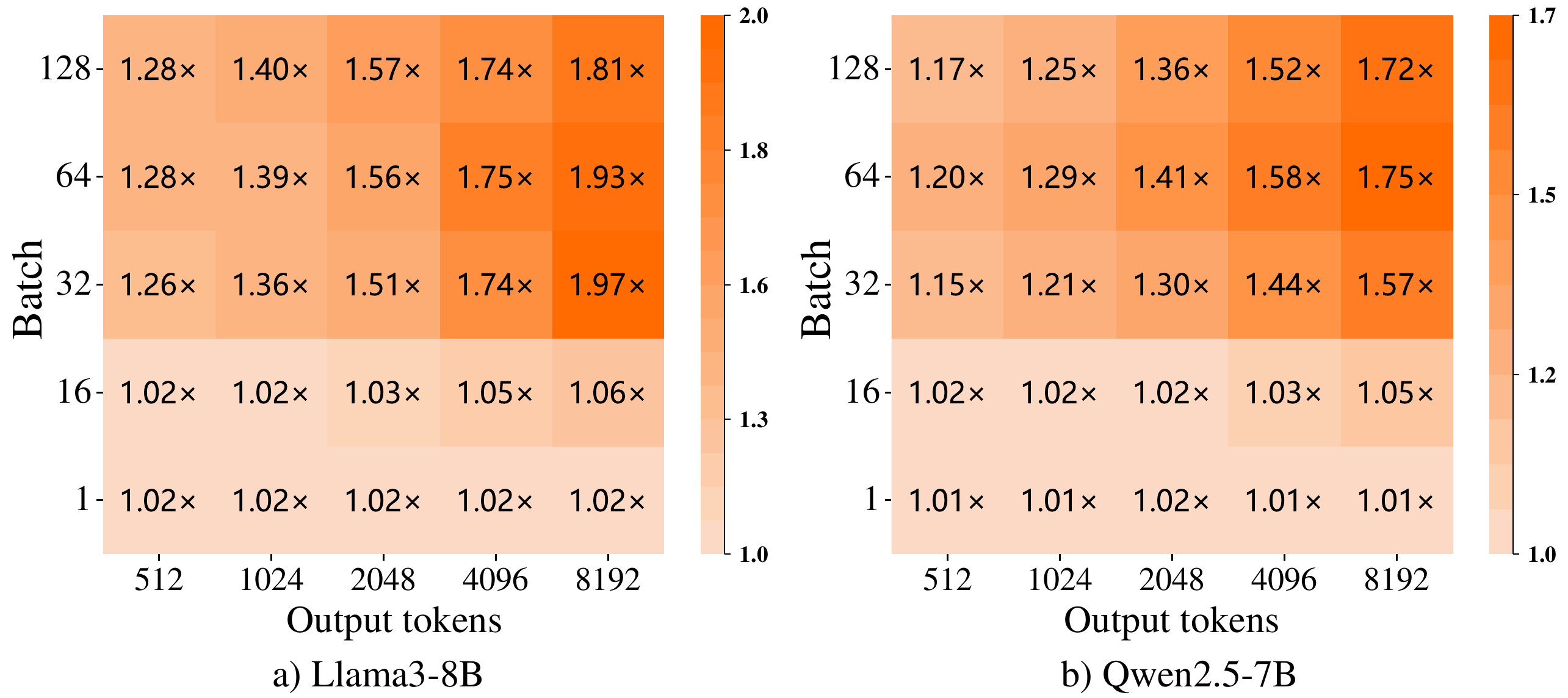}
\caption{Speedups of the proposed method over native XFormers across varying batch sizes and output sequence lengths, evaluated on a single NVIDIA H20 GPU.}
\label{fig:heatmap}
\end{figure}

The output sequence length and batch size directly impact the number of KV Caches during inference, thereby affecting the performance of the proposed method. To systematically evaluate the influence mechanisms of these arguments on method performance, we conduct experiments within the argument space spanning output tokens of 512-8192 and batch sizes of 1-128. By observing the throughput speedup ratio achieved by the proposed method compared to the native XFormers backend, we reveal the regulation patterns of performance optimization space under different argument configurations. The experimental results are fully demonstrated in Figure \ref{fig:heatmap}.

Under fixed batch size configurations, increasing the output sequence length escalates per-sequence KV Cache volume, resulting in cumulative growth of KV block loading latency in native XFormers backend. In this scenario, our methodology amplifies performance improvements through optimized KV block loading efficiency - the acceleration effect demonstrates accumulative amplification with growing block quantities, ultimately achieving monotonically increasing speedup ratios. Similarly, when maintaining constant sequence lengths, batch size expansion proportionally multiplies KV block quantities, thereby fully unleashing the architectural advantages of our proposed prefetching methodology.
Experimental results reveal that the peak speedup ratio deviates from the anticipated argument combination of 8K output tokens and batch size 128, attributable to dual hardware constraints: First, the batch size exceeds the performance acceleration boundary of L2 cache, where the required KV Block volume surpasses the cache's capacity; Second, the total KV Cache size generated by large batch-long sequence configurations transcends the GPU's physical memory limits, activating the recomputation mechanism in the vLLM framework and inducing throughput degradation.

\subsubsection{Multi-GPU Throughput}

\begin{figure}
\centering
\includegraphics[scale=0.19]{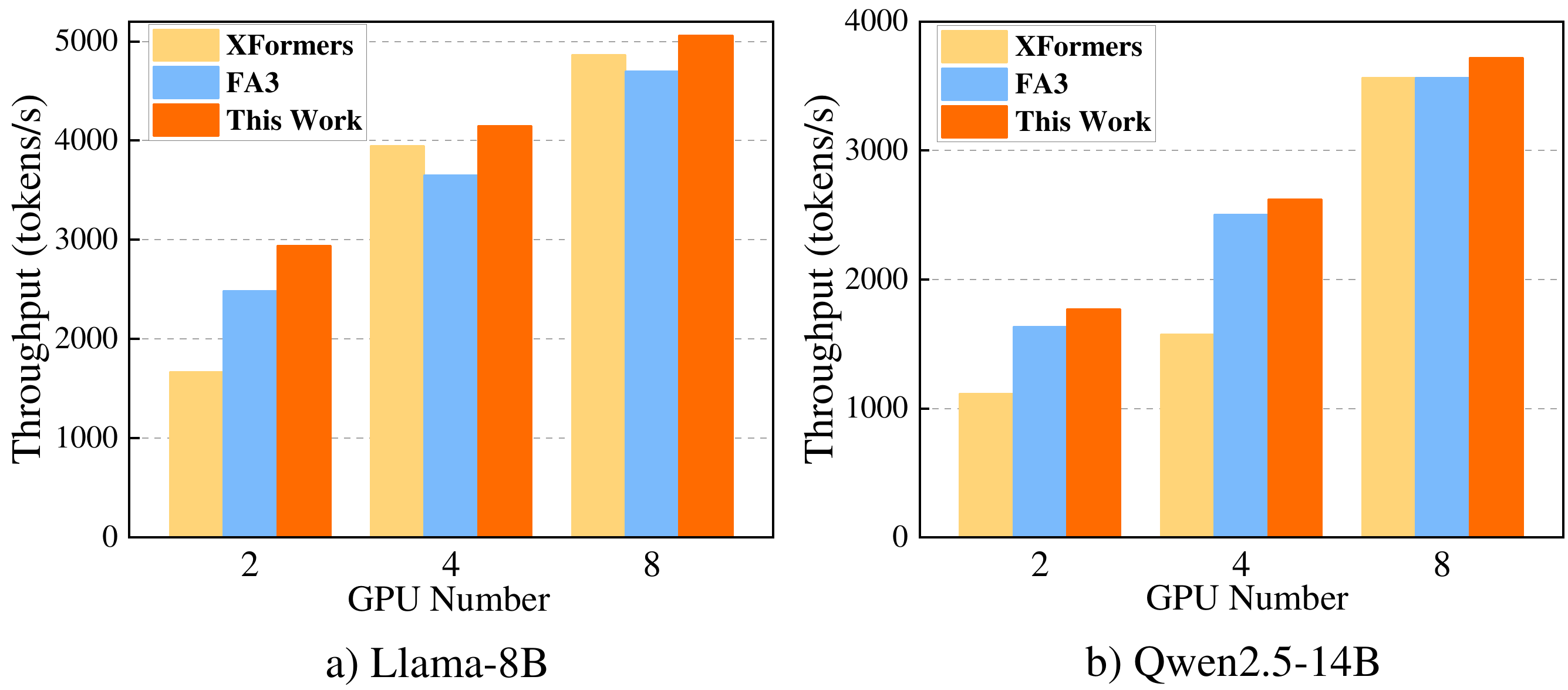}
\caption{Multi-GPU end-to-end inference throughput comparison across backends under fixed 4096 output tokens with batch size 64, benchmarked on NVIDIA H20 GPUs.}
\label{fig:multigpu}
\end{figure}

We further evaluate the performance of the proposed method under multi-GPU tensor parallelism (TP) configurations. Benchmark tests are conducted across 2/4/8 GPU setups using Llama3-8B and Qwen2.5-14B models, with fixed output sequences of 4096 tokens and a batch size of 64.
As demonstrated in Figure \ref{fig:multigpu}, the proposed method demonstrates absolute performance superiority compared to other baseline backends. Relative to the native XFormers backend, it achieves performance improvements of 4\%-59\% and 4\%-76\% on Llama3-8B and Qwen2.5-14B models, respectively. Notably, the performance gains exhibit diminishing returns with increasing tensor parallelism (TP) scale, a phenomenon originating from the attention head partitioning mechanism in TP —when scaling GPU counts from 2 to 8, the number of attention heads processed per GPU reduces to 1/2, 1/4, and 1/8 of the initial allocation, thereby concurrently diminishing the optimizable I/O operation space. Specially, the FA3 baseline exhibits 3-7\% throughput regression in 4/8-GPU configurations for Llama3-8B, whereas our method maintains acceleration through KV Block prefetching.

\section{Related Work}

\subsubsection{Attention Kernel \& KV Cache Optimization}

The autoregressive nature of the Transformer architecture restricts the inference performance of large language models (LLMs) primarily due to off-chip memory bandwidth limitations. To address this memory bottleneck, researchers have proposed solutions from two directions: computational kernel optimization and KV Cache structure refinement. In computational kernel optimization, the FlashAttention-1/2 \cite{dao2022flashattention,dao2023flashattention} series employs tiling and kernel fusion strategies to significantly reduce GPU high-bandwidth memory (HBM) access frequency, while the latest FlashAttention-3 \cite{NEURIPS2024_7ede97c3} further leverages Hopper architecture's hardware features to achieve dual improvements in computational speed and precision. DeepSpeed-inference \cite{aminabadi2022deepspeed} introduces deep fusion technology that integrates multiple operations into a single kernel and customizes GeMM kernels for small-batch inference scenarios to enhance memory bandwidth utilization. In KV Cache structure optimization, Multi-Query Attention (MQA) \cite{shazeer2019fast} and Grouped-Query Attention (GQA) \cite{Ainslie2023GQATG} mechanisms reduce KV Cache memory consumption by sharing key-value caches across multiple query heads, and Multi-Head Latent Attention (MLA) \cite{liu2024deepseek} employs low-rank joint compression techniques to minimize KV cache during inference, achieving memory footprint compression with controlled accuracy loss. It is noteworthy that the prefetching method proposed in this paper exhibits orthogonality with the aforementioned optimizations, allowing the prefetching method to integrate with existing techniques for collaborative optimization of memory bottlenecks in inference.

\subsubsection{Prefetching}

To hide GPU memory and I/O latencies, existing prefetching research includes the following: ZeRO-Infinity \cite{rajbhandari2021zero} asynchronously prefetches subsequent layer weights from CPU memory or NVMe into GPU memory during the computation of the current layer, thereby hiding host-device communication latency. DeepUM \cite{jung2023deepum} employs correlation prefetching to mask page migration latency between GPU and CPU memory. While these methods show promise for cross-device data transfers, they are unsuitable for pure GPU-based LLM inference workflows. PRESERVE \cite{Yuzuguler2025PRESERVEPM} prefetches model weights and KV Cache from off-chip memory into L2 cache during multi-GPU parallel allReduce operations, effectively masking data loading latency. However, this method is only effective in multi-GPU parallel scenarios and gradually loses efficacy as batch size increases. In contrast, the proposed method accelerates attention score computation at the CUDA kernel level, ensuring sustained performance gains across all scenarios.

\section{Conclusion}

In this paper, we propose an L2 cache-based asynchronous prefetching method for KV Cache that effectively mitigates memory bandwidth bottlenecks in large model inference. Through comprehensive experiments on mainstream models including Llama3-8B, we demonstrate the method achieves 2.15× kernel acceleration and 1.97× throughput improvement over native XFormers on H20 GPU, outperforming state-of-the-art baselines FlashAttention-3. The proposed technique maintains orthogonality with existing optimization techniques and can be integrated into current inference frameworks to form synergistic optimization combinations, providing an extensible latency-hiding solution for next-generation LLM inference engine design.


\bibliography{main}

\end{document}